\documentclass[letterpaper, 10 pt, conference]{ieeeconf}  %

\IEEEoverridecommandlockouts                              %

\usepackage{graphicx}
\usepackage[dvipsnames]{xcolor}
\usepackage{colortbl}
\usepackage{epsfig} %
\usepackage{mathptmx} %
\usepackage{times} %
\usepackage{amsmath} %
\usepackage{amssymb}  %
\usepackage[bookmarks=false,colorlinks=true,
            linkcolor=black,citecolor=black,urlcolor=black]{hyperref}
\usepackage{cleveref}
\usepackage{multirow}
\usepackage{makecell}
\usepackage{booktabs}
\usepackage{pifont}
\usepackage{cleveref}
\usepackage{hyperref}
\usepackage{color}
\usepackage[normalem]{ulem} %
\crefname{equation}{Eqn.}{equations}
\crefname{table}{Table}{tables}
\crefname{figure}{Fig.}{figures}
\crefname{section}{Sec.}{sections}

\PassOptionsToPackage{table}{xcolor}

\newcommand{\xmark}{\ding{55}}%

\definecolor{yellow}{rgb}{1, 1, 0.7}
\definecolor{orange}{rgb}{1, 0.85, 0.7}
\definecolor{red}{rgb}{1, 0.7, 0.7}
\definecolor{red0}{rgb}{1, 0., 0.}
\definecolor{red2}{rgb}{0.90, 0.17, 0.13}

\definecolor{teal}{rgb}{0.0, 0.5, 0.5}
\definecolor{goldenrod}{rgb}{0.867, 0.769, 0.255}
\definecolor{gray}{rgb}{0.843, 0.843, 0.843}
\definecolor{brown}{rgb}{0.494, 0.259, 0.0196}
\definecolor{grey}{rgb}{0.9,0.9,0.9}
\definecolor{greenx}{rgb}{0.527, 0.859, 0.796}
\definecolor{red2}{rgb}{1.0, 0.0, 0.0}

\def\methodname{SaLF}

\title{\LARGE \bf
SaLF: Sparse Local Fields for Multi-Sensor Rendering in Real-Time
}

\author{Yun Chen$^{1,2*}\thanks{*Equal contributions.}$\quad Matthew Haines$^{1,3*\dag}\thanks{\dag Work done while a research intern at Waabi.}$ \quad  Jingkang Wang$^{1,2}$ \quad Sahil Jain$^{1}$  \\  
Krzysztof Baron-Lis$^{1}$  \quad Sivabalan Manivasagam$^{1,2}$ \quad Ze Yang$^{1,2}$ \quad Raquel Urtasun$^{1,2}$ \\
\textsuperscript{1}Waabi  \quad
\textsuperscript{2} University of Toronto \quad
\textsuperscript{3}University of Waterloo \\
\texttt{\small \{ychen, jwang, sjain, klis, siva, zyang, urtasun\}@waabi.ai, m4haines@uwaterloo.ca
}
}

\begin{document}

\maketitle

\thispagestyle{empty}
\pagestyle{empty}

\begin{abstract}
    High-fidelity sensor simulation of light-based sensors such as cameras and LiDARs is critical for safe and accurate autonomy testing.
    Neural radiance field (NeRF)-based methods that reconstruct sensor observations via ray-casting of implicit representations have demonstrated accurate simulation of driving scenes, but are slow to train and render, hampering scalability.
    3D Gaussian Splatting (3DGS) has demonstrated faster training and rendering times through rasterization, but is primarily restricted to pinhole camera sensors, preventing usage
    for realistic multi-sensor autonomy evaluation.
    Moreover, both NeRF and 3DGS couple the representation with the rendering procedure (implicit networks for ray-based evaluation, particles for rasterization), preventing interoperability, which is key for general usage. In this work, we present Sparse Local Fields (SaLF), a novel volumetric representation that supports rasterization and raytracing for unified multi-sensor simulation.
    \methodname\ represents volumes as a sparse set of 3D voxel primitives, where each voxel is a local implicit field. \methodname\ has fast training ($<$30 min) and rendering capabilities
    (50+ FPS for camera and 600+ FPS for LiDAR), has adaptive pruning and densification to easily handle large scenes, and can support non-pinhole cameras and spinning LiDARs. We demonstrate that SaLF has similar realism as existing self-driving sensor simulation methods while improving efficiency and enhancing capabilities, enabling more scalable simulation.
\end{abstract}

\section{Introduction}

Closed-loop simulation is an integral part of testing self-driving vehicles~\cite{wang2021advsim,sarva2023adv3d}. To evaluate the full autonomy system, modern simulators must simulate the sensors (e.g., LiDAR, camera) that the vehicle utilizes for perception. Such multi-modal systems must be realistic to accurately measure autonomy performance, and highly efficient to enable scalable testing and training.

Neural Radiance Field (NeRF) representations~\cite{mildenhall2020nerf} have made significant progress in realistic 3D multi-sensor self-driving simulation~\cite{unisim,tonderski2024neurad}. These methods model the scene as composable dynamic actors and static backgrounds using 3D implicit representations. {By casting rays according to sensor intrinsics and extrinsics, NeRF-based methods} generate high-fidelity sensor data via volume rendering. However, high computational demands during training (hours per scene) and rendering (1-2 FPS) limit their scalability for real-time simulation, {especially given modern vehicles utilize over 20 sensors~\cite{waymo2024sixthgen}.} While several methods improve rendering efficiency by baking into meshes~\cite{chen2022mobilenerf,yariv2023bakedsdf,hedman2021baking,liu2023real} or grid look-ups~\cite{reiser2023merf,yu2021plenoctrees}, they still suffer from time-consuming training, complex baking procedures, and potential quality degradation compared to the original representation.

\begin{figure}[t]
	\centering
    \includegraphics[width=0.88\linewidth]{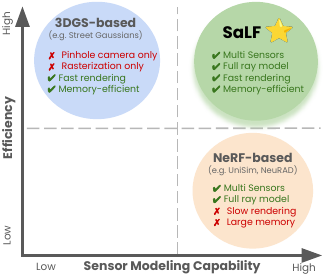}
	\vspace{-0.1in}
    \caption{\textbf{SaLF combines high efficiency with advanced sensor modeling capabilities for self-driving simulations.}
	}
	\vspace{-0.1in}
	\label{fig:teaser_speed}
\end{figure}

\begin{figure*}
    \centering
    \includegraphics[width=0.83\textwidth]{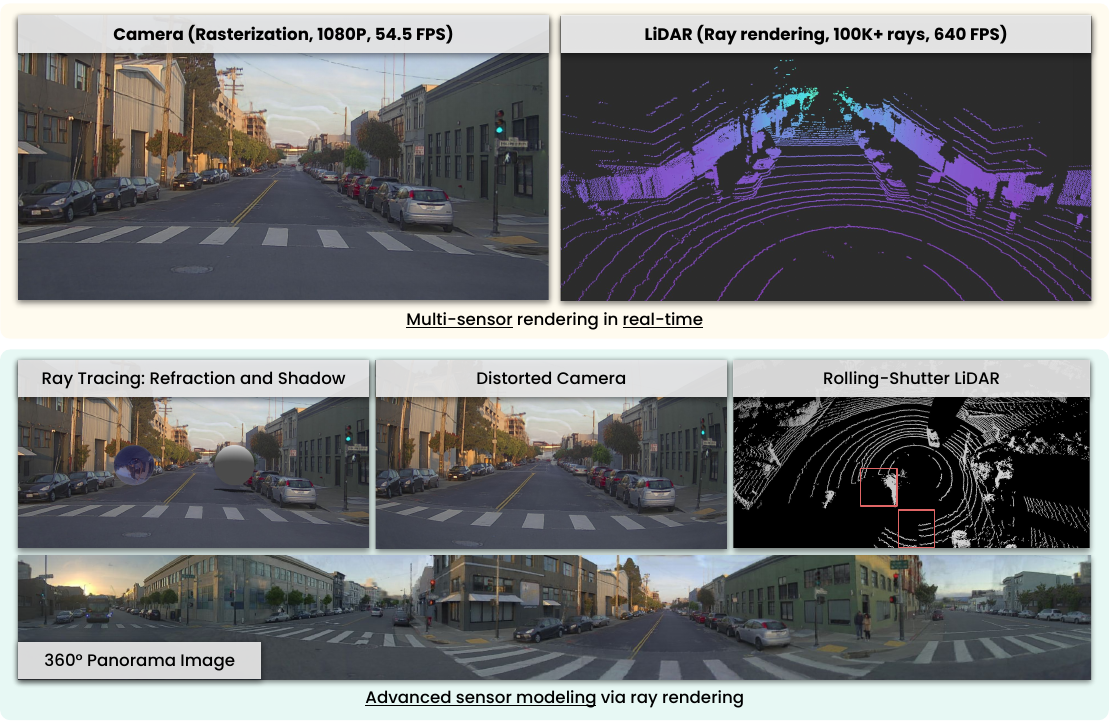}
	\vspace{-0.1in}
    \caption{
        \textbf{Real-time self-driving sensor simulation with \methodname{} representation.}
        Our method achieves high-performance rendering for both camera and LiDAR,
        and supports advanced features
        including {secondary effects (e.g., refraction, reflection, and shadow) and complex sensor models (e.g., fisheye, rolling-shutter, and panoramic cameras)}.
        {This is made possible by an}
        efficient and unified representation {that supports} both
        rasterization and ray-tracing.
    }
	\vspace{-0.05in}
    \label{fig:teaser}
\end{figure*}

3D Gaussian Splatting (3DGS)~\cite{3dgs,yan2024street,zhou2024drivinggaussian} enables fast training and real-time pinhole camera rendering by modeling scenes as explicit 3D Gaussian particles and rasterizing them onto the image plane. However, like other rasterization-based approaches, 3DGS lacks native support for ``ray-based'' rendering, which is required for complex sensor models such as rolling-shutter LiDARs or fish-eye cameras. It also struggles to accurately model phenomena like motion blur and secondary lighting effects (e.g., {refraction}), which are crucial for robust testing~\cite{manivasagam2023towards,pun2023lightsim}. These limitations restrict its use in comprehensive testing despite its real-time capabilities. {Furthermore, similar to how graphics meshes support both raytracing and rasterization, we argue for a unified, learnable volumetric representation.} {While specialized LiDAR or camera models may achieve high performance in their respective domains, maintaining disjoint representations for the same scene adds prohibitive memory overhead and pipeline complexity. Furthermore, a shared volumetric foundation is essential to guarantee physical consistency across simulated sensors, ensuring that downstream autonomy systems do not fail due to misalignment.}

{In this work, we present Sparse Local Fields (SaLF), a volumetric representation supporting both rasterization and raytracing for multi-sensor simulation.} \methodname\ consists of voxel primitives, each acting as a local implicit field mapping spatial coordinates to geometry and appearance. Like NeRF, \methodname\ supports volume rendering via ray-casting {to accurately model} LiDAR and complex cameras. Without any baking, \methodname's\ voxels are natively backed by an octree, directly supporting accelerated raytracing. Like 3DGS, \methodname\ can be {efficiently} rasterized for fast pinhole camera rendering. \methodname\ also supports adaptive voxel pruning and densification, creating compact representations for large scenes.

By unifying support for diverse sensor models and complex phenomena such as rolling-shutter effects and refraction, SaLF is highly valuable for scalable autonomous driving simulation. Experiments on a public self-driving dataset demonstrate that \methodname\ achieves comparable realism to prior works while being significantly more efficient to train and render. {We showcase its raytracing capabilities for fast and {versatile} sensor simulation,} and demonstrate that \methodname{} achieves a smaller domain gap for downstream perception, prediction, and planning tasks.

\section{Related Work}

\paragraph{Efficient NeRFs} Neural Radiance Fields (NeRF)~\cite{mildenhall2020nerf} provide a foundation for photorealistic 3D reconstruction, yet the vanilla formulation remains computationally intensive. Recent works address these efficiency challenges. DVGO~\cite{sun2021direct} replaces the global MLP with a sparse 3D grid for faster convergence, while Plenoxels~\cite{yu2021plenoxels} utilizes explicit spherical harmonics in a sparse grid, removing neural networks entirely. Instant-NGP~\cite{mueller2022instant} further adopts multi-resolution hash encoding for {compactness.} To enable real-time rendering, ``baking'' methods pre-compute neural field properties into efficient representations like sparse voxel grids~\cite{hedman2021baking}, {triplanes}~\cite{reiser2023merf}, octrees~\cite{yu2021plenoctrees}, or VDB~\cite{garbin2021fastnerf}. Other works extract explicit meshes~\cite{chen2022mobilenerf,yariv2023bakedsdf,liu2023real} for standard rasterization pipelines. However, these approaches often face intractable memory usage in large scenes, quality degradation when baked~\cite{liu2023real, garbin2021fastnerf,yu2021plenoctrees}, or long training times~\cite{hedman2021baking,chen2022mobilenerf,yariv2023bakedsdf}. SVR~\cite{sun2024sparse} performs voxel rasterization but focuses strictly on cameras without supporting ray-based sensor models. In contrast, \methodname{} optimizes and renders efficiently for multi-sensor simulation in large scenes without baking.

\paragraph{3DGS} 3D Gaussian Splatting (3DGS)~\cite{3dgs} enables real-time photorealistic rendering by representing scenes with oriented 3D Gaussians and tile-based rasterization. {3DGS has been applied to} 3D generation~\cite{yi2024gaussiandreamer} and camera simulation~\cite{zhou2024drivinggaussian,yan2024street,chen2024g3r, wang2025flux4d}. However, like other rasterization-based approaches, 3DGS assumes a pinhole camera model and lacks the flexible ray-based rendering of NeRF, limiting its use for rolling-shutter LiDARs and fish-eye cameras. Several concurrent works~\cite{wu20243dgut, chen2024lidar, ren2024unigaussian, kung2024lihi, hess2024splatad, liao2024fisheye} have extended 3DGS to non-pinhole sensors by approximating sensor-specific effects, but these solutions often hinder generalization. Concurrent works enable 3DGS ray tracing by wrapping Gaussians in proxy geometries for BVH traversal ~\cite{moenne-loccoz2024gaussian, zhou2024lidar}. This introduces significant structural overhead compared to SaLF, which natively supports both paradigms within a unified representation.

\paragraph{{Data-driven} Sensor Simulation for Self-Driving} {Traditional graphics-based simulators~\cite{dosovitskiy2017carla,shah2018airsim} face significant domain gaps in geometry and appearance.} Data-driven {neural rendering}~\cite{ost2021neural,unisim} has attracted attention for its photorealism and ability to reconstruct real-world scenes. Methods like~\cite{ost2021neural,unisim,tonderski2024neurad,yang2025genassets} use NeRFs to build digital twins, decomposing scenes into backgrounds and actors to enable realistic camera {and LiDAR \cite{unisim, wu2024dynamic,tonderski2024neurad}} simulation. However, these typically require hours of training per 10-second clip and cannot perform {real-time} rendering. To address this, recent works~\cite{yan2024street,zhou2024drivinggaussian,chen2024omnire} leverage compositional 3DGS for real-time camera simulation {but remain restricted to pinhole models.} In contrast, we build the first self-driving {neural} sensor simulator supporting real-time rendering for {complex} cameras and LiDAR.

\section{\methodname\: Sparse Local Fields}

We present \methodname\ (Sparse Local Fields), a novel volumetric representation that supports efficient tile-based rasterization and {flexible}, high-fidelity ray-casting of complex scenes.
In this section, we detail our scene representation (\cref{sec:rep}) and our rasterization and ray-casting rendering algorithms (\cref{sec:rendering}), the coarse-to-fine densification strategy {for compactness and efficient training} (\cref{sec:adaptive}), and discuss how \methodname\ compares to {NeRF and 3DGS} (\cref{sec:comparison}).

\subsection{Representation}
\label{sec:rep}

\methodname\ represents scenes using a sparse grid of local implicit fields {that map} global 3D coordinates and view-directions to {spatial} properties such as density and color.
Let $\mathcal{V} \subset \mathbb{R}^3$ be a three-dimensional volume {in}
an axis-aligned bounding box (AABB) with dimensions $(V_h, V_w, V_d)$.
We partition $\mathcal{V}$ into a regular grid $\mathcal{G}$ of dimensions $\lceil V_h/s_0 \rceil \times \lceil V_w/s_0 \rceil \times \lceil V_d/s_0 \rceil$,
where each voxel is a {cube} with edge length $s_0$.
{Each voxel supports recursive sub-division into 8 smaller voxels, up to $K$ levels ($s_0 \dots s_k$).}
To efficiently handle large-scale scenes, we employ a sparse representation that stores only non-empty voxels.

{
Each voxel is characterized by its static geometric parameters: position $p \in \mathbb{R}^3$,  {scale} $s \in \mathbb{R}$, and rotation $q \in \mathbb{R}^4$. 
$q$
represents the orientation relative to the global coordinate system, with all voxels initialized to the identity quaternion in the global frame.
}
Each voxel also contains
a geometry field $W_\sigma \in \mathbb{R}^{1 \times 4}$ and a color field $W_c \in \mathbb{R}^{3 \times 3}$, {along with} 2nd order spherical harmonics $W_{sh} \in \mathbb{R}^{3 \times 4}$ for view-dependent lighting effects.
For any point inside the voxel, let {$\textbf{x} \in [-1,1]^3$} denote its normalized local coordinates, and $\hat{\textbf{x}} = [\textbf{x}, 1]$ be its homogeneous representation.
The geometry field $f_\sigma(\textbf{x}; W_\sigma): [-1,1]^3 \rightarrow \mathbb{R}_+$ computes the density as:
\begin{equation}
\sigma = f_\sigma(\textbf{x}; W_\sigma) = \mathrm{exp}({W_\sigma \hat{\textbf{x}}^T}).
\label{eq:density}
\end{equation}
{Similarly, the color field
$f_c(\textbf{x},\mathbf{\omega}; W_c,W_{\text{sh}}): [-1,1]^3 \times \mathbb{S}^2 \rightarrow [0,1]^3$
computes the color value as:
\begin{equation}
    c = \mathrm{sigmoid} (W_c {{\textbf{x}}}^T + W_{\text{sh}}\gamma(\mathbf{\omega}))
\label{eq:color}
\end{equation}
where $\mathbf{\omega} \in \mathbb{S}^2$ is the view direction and $\gamma: \mathbb{S}^2 \rightarrow \mathbb{R}^{4}$ maps the view direction to spherical harmonics basis coefficients.
}
\begin{figure}[tb]
    \centering
\includegraphics[width=0.99\linewidth]
               {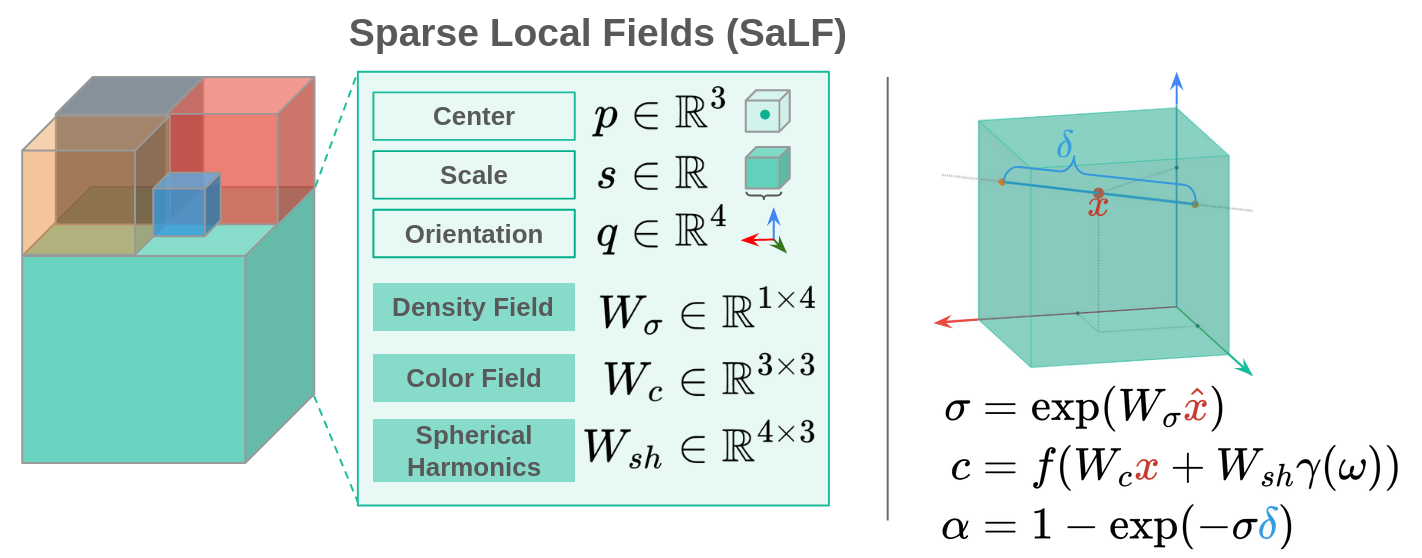}
	\vspace{-0.2in}
               \caption{\textbf{\methodname\ Representation.}
               \textbf{Left:}
               \methodname\ models scenes using an adaptive sparse voxel grid with variable scales.
               Each voxel is characterized by$\fcolorbox{greenx}{white}{static parameters}$and$\colorbox{greenx}{learnable parameters}$.
            \textbf{Right:} %
           Within a voxel, for any point with normalized coordinates $\textcolor{red0}{\textbf{x}}$, the density  $\sigma$ and color $c$ values are derived from $W_\sigma$ and $W_c$ along with the encoded view direction $\gamma(\omega)$ modulating $W_{\text{sh}}$.
       The opacity $\alpha$ of a ray is calculated using the density at the intersection midpoint and the traversal distance $\textcolor{blue}{\delta}$.}
\label{fig:repr}
\end{figure}

\paragraph{Volume rendering:}
We render the color $\hat{C}$ for each ray by accumulating the color and opacity values of intersected voxels along the ray, following the volume rendering equation as in NeRF:
\begin{equation}
\hat{C} = \sum_{i=1}^{N_c} T_i\alpha_i c_i,
\label{eq:volumerendering}
\end{equation}
where $N_c$ is the number of intersected voxels along the ray, $T_i$ represents the accumulated transmittance $T_i = \prod_{j=1}^{i-1} (1 - \alpha_j)$
and $\alpha_i$ denotes the opacity computed from the density value $\sigma_i$ and ray segment length $\delta_i$ within the $i$-th voxel:
\begin{equation}
\alpha_i = 1 - \exp(-\sigma_i \delta_i).
\label{eq:opacity}
\end{equation}
The  density $\sigma_i$ and color $c_i$ values are sampled at the midpoint of each ray-voxel intersection segment using \cref{eq:density} and \cref{eq:color}, respectively.

\paragraph{Surface parameterization}
Autonomous driving scenes require high-quality surfaces to {enable} accurate LiDAR simulation and secondary ray effects such as reflection.
Instead of directly representing density {as $W_\sigma$}, we follow previous approaches~\cite{unisim,tonderski2024neurad}  to adopt a signed distance function (SDF) {for better surface parameterization}. %
The {geometry SDF field}
denoted as $W_s$, quantifies the signed distance ($s_\pm$) between a point $\textbf{x}$ and the surface as:
\begin{equation}
s_{\pm} = W_s\hat{\textbf{x}}^T
\label{eq:sdf}
\end{equation}
We transform $s_\pm$ to density $\sigma$ following VolSDF~\cite{yariv2021volsdf, siddiqui2024meta}:
\begin{equation}
{\sigma} = \frac{a}{2} + \frac{a}{2} \text{sign}(s_{\pm})\left(1 - e^{-|s_{\pm}|/b}\right).
\label{eq:sdf2sigma}
\end{equation}
{The final learnable parameters of each voxel are the geometry field $W_s$, the color field $W_c$, the spherical harmonics $W_{sh}$, and the SDF-to-density parameters $a, b$.}

\subsection{Efficient Rendering of \methodname}%
\begin{figure}[tb]
	\centering
\includegraphics[width=0.99\linewidth]
               {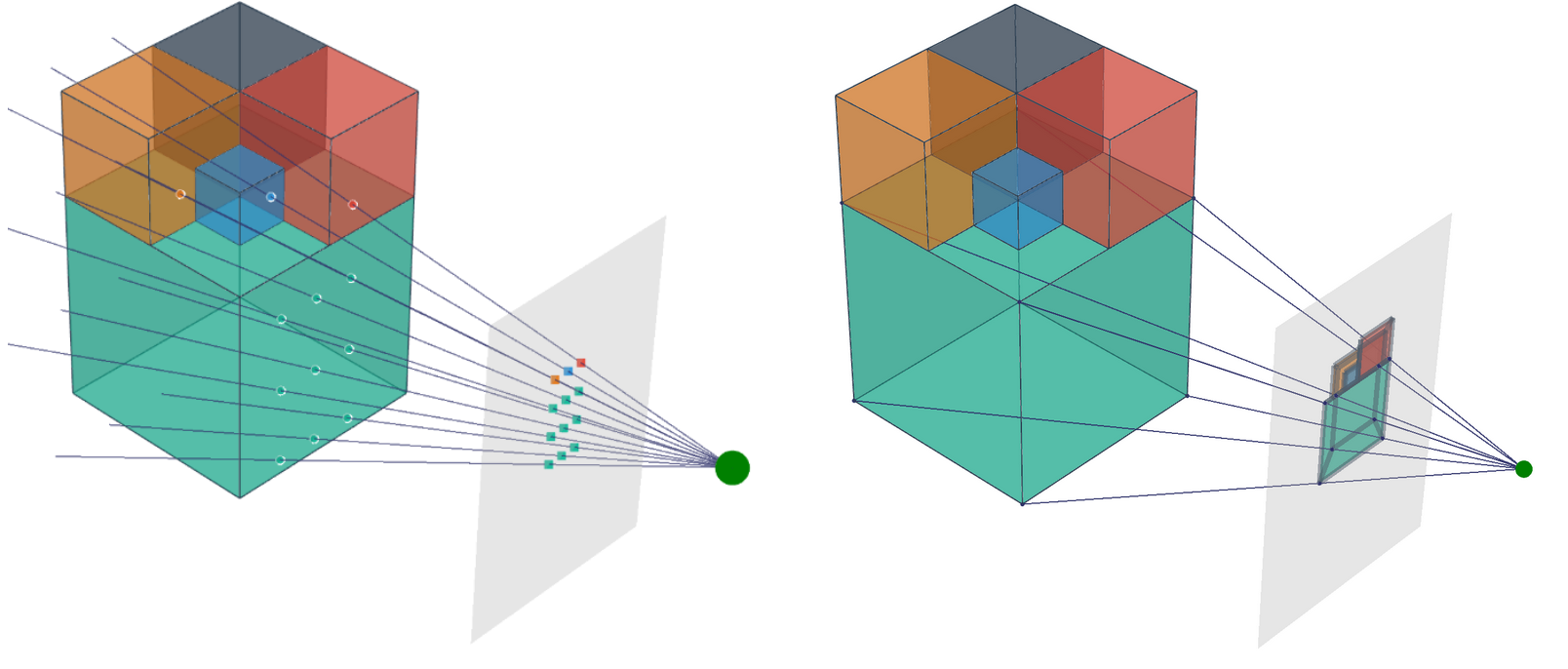}
               \vspace{-0.2in}
               \caption{\textbf{
                   \methodname \ can be efficiently rendered by
           {ray casting} (left) and {rasterization} (right).} %
           Ray casting is more flexible and can handle more complex physics phenomena, while {rasterization} is more efficient for pinhole cameras. Both follow the same rendering equation. }
	\label{fig:rendering}
\end{figure}

\label{sec:rendering}
As shown in \cref{fig:rendering}, \methodname{} is interoperable and can be rendered {via ray-{casting} through the volume and computing ray-voxel intersections,}
or by {splatting and compositing}
voxels onto the image plane (i.e., {rasterization}).
Both can be implemented following the volume rendering equation in \cref{eq:volumerendering}, {with rasterization being fast{er}
  but making more approximations in the image formation process}. %
The ray-casting approach is more flexible and can handle more complex physics phenomena {and sensor models}.
{We now discuss in more detail our efficient implementations of each.}

\paragraph{Ray-{Casting} with Octree Acceleration}
\methodname\ can be rendered as NeRF by casting rays through the voxels from the sensor origin, sampling points, and accumulating their color and opacity.
To accelerate ray-marching and ray-voxel intersection checks, we employ an octree data structure. The octree recursively partitions the volume into eight sub-volumes, where non-leaf nodes maintain pointers to their sub-volumes, and leaf nodes either store $-1$ for empty space or a pointer to the corresponding voxel.
This hierarchical structure enables fast ray traversal through the volume.
Through a single ray-box intersection test, empty regions can be bypassed.
 Upon encountering non-empty nodes, traversal selectively descends into intersected child octants with logarithmic complexity.
\paragraph{Tile-based Splatting}
\methodname\ can also be rendered by
splatting, which projects voxels onto the image plane, compositing them with alpha blending.
Following 3DGS, we divide the image plane into a grid of $16\times 16$ tiles.
For each tile, we perform view-frustum culling to identify and sort relevant voxels.
Each tile is rendered by a thread block to iterate over the relevant voxels.
Crucially, we preload these voxels into shared memory to reduce global memory access, which is a key optimization that significantly accelerates rendering.
The opacity for each voxel is computed based on current pixel's ray travel distance within each voxel using \cref{eq:opacity}, and the color is sampled from the color field at the intersection point.
{For a pinhole camera with primary rays,} 
the splatting achieves significant acceleration through tile-based processing and shared memory.
\subsection{Initialization and Densification}
\label{sec:adaptive}
\begin{figure}[tb]
	\centering
\includegraphics[width=0.99\linewidth]
               {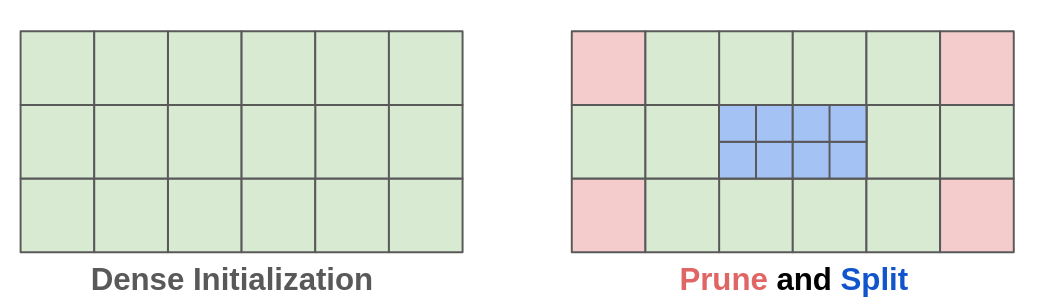}
               	\vspace{-0.2in}
               \caption{\textbf{Initialization and Densification.} \methodname \ initializes the scene representation with a coarse regular grid partitioning, then adaptively prune \textcolor{red2}{empty region} while \textcolor{blue}{densifying regions}  that need fine-details.}
               	\vspace{-0.1in}
	\label{fig:densification}
\end{figure}

Naive uniform voxelization {of large-scale scenes} at high resolution leads to prohibitive memory consumption that scales cubically with scene size.
To address this challenge, we adopt a coarse-to-fine approach as shown in \cref{fig:densification}. {During training, we} initialize the scene with a
coarse representation {and apply an adaptive densification and pruning strategy.}
Voxels exhibiting significant color field gradients {(evaluated as the $L_1$ norm of the loss gradient with respect to the color field parameters $W_c$)} are subdivided into eight child voxels in an octree-aligned manner.
{Voxels with negligible opacity {($\alpha < 0.001$)} are removed from the sparse set.
Densification and pruning together enables preserving of fine details while maintaining a compact memory footprint.}

\subsection{Comparison with 3DGS and NeRF}
\label{sec:comparison}
\methodname\ represents {implicit} scenes using discrete volumes like other voxel-based NeRF variants such as DVGO \cite{sun2021direct} and Plenoxels \cite{yu2021plenoxels}.
Unlike these methods, whose uniform grids cause an $O(N^3)$ memory explosion, our approach ensures memory efficiency at scale by allocating finer voxels exclusively to complex geometry via gradient-based octree subdivision.

Both \methodname\ and 3DGS perform efficient splatting through sparse scene representations.
But 3DGS does not support ray-based rendering {directly} because multiple Gaussians can contribute to the same point when {overlapped}.
In contrast, \methodname\ defines distinct implicit functions for non-overlapping regions, enabling straightforward ray-voxel intersection and {property} evaluation.
Furthermore, 3DGS often struggles with {limited} surface quality due to disconnected semi-transparent Gaussian primitives and {often requires significant regularization~\cite{cheng2024gaussianpro,huang20242d}}, while \methodname\ can leverage established NeRF techniques {such as SDF} for surface reconstruction.
The initialization and densification strategies also differ: 3DGS requires sparse points as initialization and {prunes, splits or clones} existing Gaussians to cover both spatial extent and fine details.
{In contrast, \methodname\ is initialized coarsely yet densely, and its hierarchically subdividing voxels automatically capture fine details while maintaining spatial coverage.}

\section{Self-driving Sensor Simulation with \methodname}
\methodname's efficient and versatile rendering capabilities are particularly well-suited for self-driving sensor simulation, which involves large scenes and dynamic actors, while demanding real-time rendering and multi-sensor support.
We now present how to utilize SaLF to construct a lightweight simulator  that achieves real-time rendering for camera and LiDAR sensors.

\subsection{Compositional Scene Representation}

\paragraph{Dynamic Scene Modelling}
Following previous work \cite{ost2021neural}, we model dynamic actors and backgrounds as distinct bounding volumes that we compose into a global frame for rendering.
For dynamic actors, we initialize voxel sets by partitioning each actor's canonical bounding box.
For rasterization,
we use actor labels to transform these dynamic voxels to the global coordinate system at each timestamp, combine them with the static voxels, and project all voxels to the image plane.
For ray-casting, we construct a separate octree for each dynamic actor.
We precompute ray-box intersections between rays and the bounding boxes of dynamic actors, sorting the entry and exit points by camera distance.
These precomputed intersections allow us to determine the traversal order between the static scene's octree and the dynamic actors' octrees, enabling efficient
ray marching.

\paragraph{Multi-scale Static Scene Initialization}

The outdoor driving environment necessitates efficient representation of large-scale scenes (e.g., sky, far-away buildings).
Our initialization strategy begins by {identifying}
a core region of
interest {that the self-driving vehicle will traverse}, which we discretize at a base resolution.
Recognizing that distant scene elements do not require high-resolution {voxels}, {we surround the core static foreground region }
with increasingly coarser outer regions.
These outer regions extend at 2$\times$, 4$\times$, 8$\times$, and 16$\times$ the base volume, with their voxel sizes scaling proportionally, naturally matching the diminishing detail requirements of distant scene elements.
We leverage LiDAR point clouds to optimize scene initialization through a three-step process: inner region voxel pruning based on point absence, subdivision of point-containing voxels, and high-opacity initialization for point-occupied voxels.

\begin{table*}[t]
	\caption{\textbf{Comparison to SoTA sensor simulation methods.} Our approach achieves real-time ($> 30$ FPS) camera and LiDAR rendering, and accelerates the reconstruction process by at least
		5× while achieving comparable realism compared to baselines. 
        Street Gaussian leverages 3DGS and does not support LiDAR simulation (\textcolor{red0}{\xmark}).
        NeuRAD leverages NeRF and adopts additional CNN decoder for higher-quality.
		We highlight \colorbox{red}{\strut first}, \colorbox{orange}{\strut second}, \colorbox{yellow}{\strut third}.
	}
		\vspace{-0.1in}
\centering
\resizebox{1.0\textwidth}{!}{
	\setlength{\tabcolsep}{11pt}
	{
		\renewcommand{\arraystretch}{1.1}

		\begin{tabular}{@{}l|cc|c|cccc}
		\toprule
		\ \multirow{2}{*}{Models} & \multicolumn{2}{c|}{Rendering FPS $\uparrow$} & \multirow{1}{*}{Recon Time $\downarrow$} & \multicolumn{4}{c}{Rendering Realism} \\
		& Camera (SaLF-Raster) & LiDAR (SaLF-Ray) & RTX-3090 hour & PSNR$\uparrow$ & SSIM$\uparrow$ & LPIPS$\downarrow$ & LiDAR-L1$\downarrow$ \\ \midrule
		\ Street Gaussian~\cite{yan2024street} & \cellcolor{red}\textbf{115.5} & \textcolor{red0}{\xmark} & 2.26 & 25.65\cellcolor{yellow} & \cellcolor{red}\textbf{0.777} & 0.307\cellcolor{yellow} & \textcolor{red0}{\xmark} \\
		\ UniSim~\cite{unisim}& 1.3 & 11.8\cellcolor{yellow} & 1.67\cellcolor{yellow} & 25.63 & 0.745 & \cellcolor{red}\textbf{0.288} & 0.100\cellcolor{orange} \\
		\ NeuRAD~\cite{tonderski2024neurad} (2x) & 1.7 & 3.79 & 3.48 & \cellcolor{red}\textbf{26.60} & 0.770\cellcolor{orange} & 0.297\cellcolor{orange} & \cellcolor{red}\textbf{0.085} \\

		\ \methodname{} (base) & 54.5\cellcolor{orange} & \cellcolor{red}\textbf{640} & 0.31\cellcolor{red} & 25.48 & 0.744 & 0.373 & 0.142 \\
		\ \methodname{} (large) & 34.3\cellcolor{yellow} & 430\cellcolor{orange} & \cellcolor{orange}0.48 & 25.78\cellcolor{orange} & 0.762\cellcolor{yellow} & 0.344 & 0.111\cellcolor{yellow} \\\hline
		\end{tabular}
	}
}
\label{tab:main_table}
\end{table*}

\subsection{Learning}
The scene representation is optimized through {the following loss function}:
\begin{equation}
\mathcal{L} = \mathcal{L}_{\text{color}} + \lambda_{\text{1}}\mathcal{L}_{\text{depth}} + \lambda_{\text{2}}\mathcal{L}_{\text{reg}},
\label{eq:loss}
\end{equation}
{where  $\mathcal{L}_{\text{color}}$ and $\mathcal{L}_{\text{depth}}$ measure} the $\ell_1$ distance between rendered and ground-truth images {and LiDAR depth, respectively}.
{$\mathcal{L}_{reg}$ enforces spatial consistency by summing the penalized differences in local SDF $s$, opacity $\alpha$, and surface normals $\mathbf{n}$ across all adjacent voxel pairs $(i, j) \in \mathcal{N}$:}
{$$\mathcal{L}_{reg} = \sum_{(i, j) \in \mathcal{N}} \left( \lambda_{s} |s_i - s_j| + \lambda_{\alpha} |\alpha_i - \alpha_j| + \lambda_{n} (1 - \mathbf{n}_i \cdot \mathbf{n}_j) \right).$$}

\section{Experiments}
\label{sec:experiments}

\subsection{Experimental Setup}
\paragraph{Dataset and Evaluation Protocol}
We evaluate our method on the PandaSet dataset~\cite{xiao2021pandaset}, which consists of 103 driving scenes captured at $1920\times1080$ resolution, with 80 frames per scene and 360-degree LiDAR data. Following previous works~\cite{unisim,tonderski2024neurad}, we use the standard 10-log evaluation set, splitting the 80 frames per log into even (training) and odd (evaluation) frames. This 50\%/50\% split is more challenging than the typical 90\%/10\% setting in other methods~\cite{zhou2024drivinggaussian}.
{We report photorealism on the front camera via} PSNR, SSIM and LPIPS (VGG-backbone)~\cite{zhang2018unreasonable} metrics. For LiDAR evaluation, we measure the median distance error ($L_1$) between the predicted and ground-truth depth\footnote[1]{NeuRAD \cite{tonderski2024neurad} reports LPIPS(AlexNet) and median $L_2$ LiDAR error.}. The training and rendering speed are reported on an RTX 3090 {averaged over the evaluation set.}
{For clarity during evaluation, we explicitly divide our unified representation into two inference modes: \textbf{SaLF-Raster} for real-time pinhole rendering, and \textbf{SaLF-Ray} (ray casting) for complex sensors (e.g., fisheye, LiDAR).}

\paragraph{Implementation Details of \methodname{}}
We implement efficient rasterization and ray-casting operations using \texttt{Taichi}~\cite{hu2019taichi, sun2023taichi3dgs}. For training, we employ the Adam optimizer with an initial learning rate of $0.01$ and apply a decay factor of 0.8 every 800 iterations, for a total of 3200 iterations. The maximum number of voxels allowed is 2.5 million. A high-capacity ``\methodname~(large)'' is also trained (5M voxels, 4500 iterations) for better performance.

\begin{figure*}[ht]
	\centering
		\vspace{-0.05in}
	\includegraphics[width=1.0\linewidth]{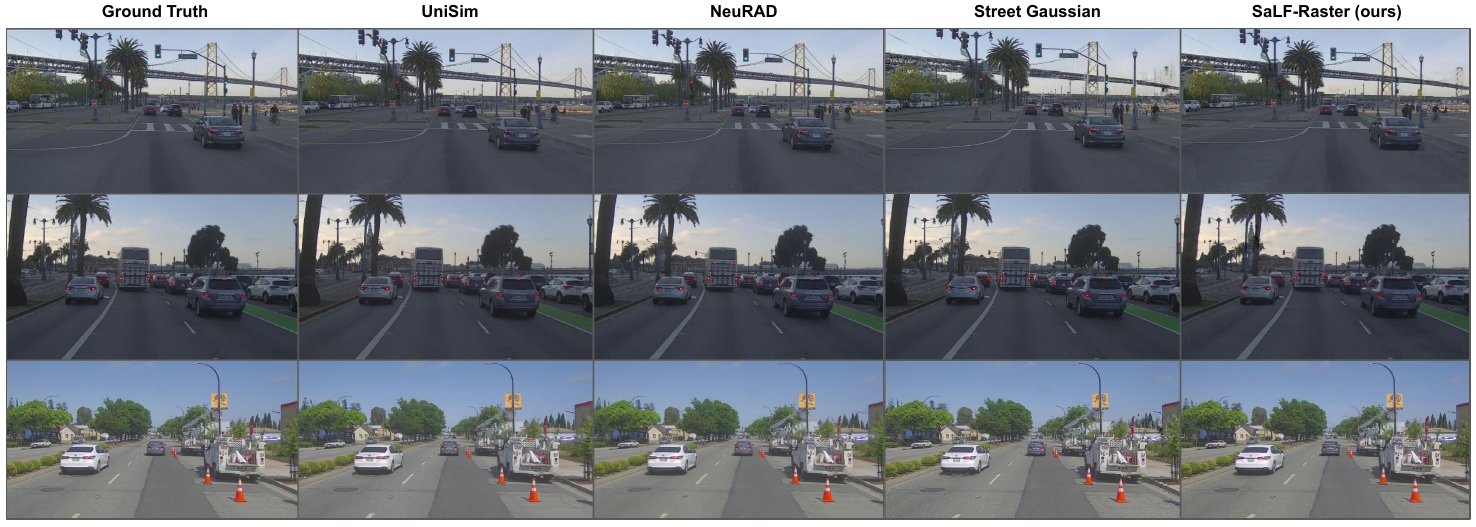}
		\vspace{-0.25in}
	\caption{\textbf{Qualitative comparison on camera novel view synthesis.} We achieve comparable photorealism compared to SoTA approaches.}
	\label{fig:qualitative}
\end{figure*}

\begin{figure}[ht]
	\centering
		\vspace{-0.1in}
	\includegraphics[width=1.0\linewidth]{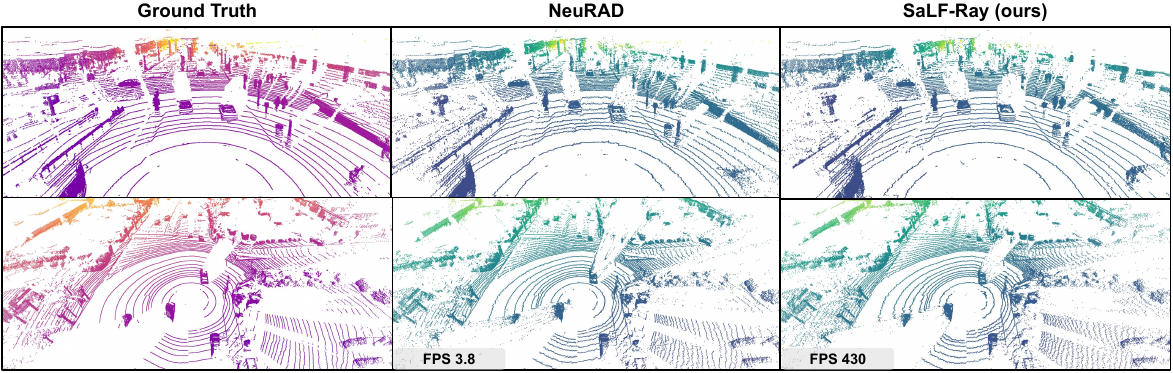}
	\caption{\textbf{Qualitative comparison on LiDAR novel view synthesis.} Our method achieves comparable LiDAR rendering performance compared to NeuRAD, while being \textit{100 $\times$ faster} in rendering.}
		\vspace{-0.1in}
	\label{fig:qualitative_lidar}
\end{figure}

\paragraph{Comparison with state-of-the-art (SoTA) methods}
We compare our approach against several SoTA methods in self-driving sensor simulation including UniSim~\cite{unisim}, NeuRAD~\cite{tonderski2024neurad} and Street Gaussian~\cite{zhou2024drivinggaussian}. UniSim leverages compositional neural feature fields to model dynamic scenes for controllable camera and LiDAR simulation. NeuRAD further extends it to handle more complex sensor phenomena (e.g., {anti-aliasing, } {rolling-shutter}, ray-dropping). Street Gaussian replaces NeRFs with compositional 3DGS and achieves real-time camera simulation, {but does not support LiDAR}.

\subsection{Fast and Realistic Multi-Sensor Simulation}
\begin{figure}[t]
	\centering
	\vspace{-0.15in}
	\includegraphics[width=1.0\linewidth]{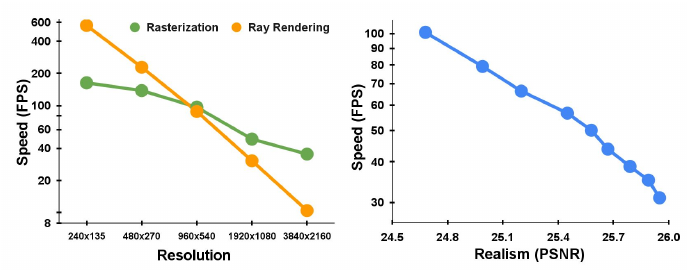}
	\vspace{-0.3in}
    \caption{\textbf{Efficiency.}
		Left: rendering speed of rasterization (SaLF-Raster) \textit{vs.} ray-tracing (SaLF-Ray) across resolutions.
		Right: trade-off between rendering speed and realism.}
	\vspace{-0.05in}
	\label{fig:efficiency-tradeoff}
\end{figure}

\begin{figure}[t]
	\includegraphics[width=1.0\linewidth]{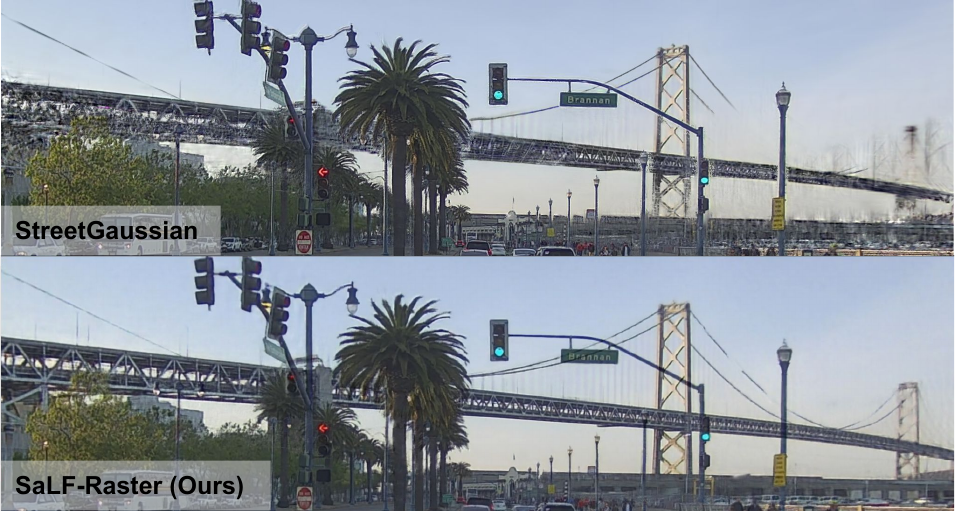}
	\vspace{-0.2in}
	\caption{\textbf{Comparison with StreetGaussian on distant regions.}
		Our method provides more accurate reconstructions for distant regions (e.g. bridge) where LiDAR and SfM points are not {sufficient.}}
	\vspace{-0.07in}
	\label{fig:init}
\end{figure}

\paragraph{Comparison to baseline methods}
We compare SoTA sensor simulation approaches on PandaSet, as shown in \cref{tab:main_table}. 
Through \textbf{SaLF-Raster}, our method achieves real-time rendering for both camera and LiDAR with comparable fidelity to SoTA simulators, and accelerates reconstruction speed by at least 5$\times$ compared to NeRF based methods. Simultaneously, \textbf{SaLF-Ray} extends this shared volumetric representation to support complex, ray-based sensor models where physical accuracy is important.
NeuRAD achieves slightly higher camera realism due to a post-processing CNN. However, without this CNN, SaLF achieves similar realism while being significantly faster, demonstrating the efficiency of our sparse voxel representation (\cref{fig:qualitative}).
{We do note that the baselines have better visual quality for dynamic actors, potentially due to their actor pose optimization during training.}
For LiDAR, \methodname\ {only has 2.5cm higher median error compared to NeuRAD, while being \emph{100$\times$ faster}. \cref{fig:qualitative_lidar} shows qualitatively similar point clouds w.r.t. the ground-truth.} 
Furthermore, our method's multi-scale initialization of coarse voxels reconstructs distant regions more accurately than 3DGS-based methods like Street Gaussian (\cref{fig:init}), which struggle due to their reliance on sparse Structure-from-Motion and LiDAR initialization.
\paragraph{Extrapolation}
To further evaluate robustness for novel view synthesis, we assess extrapolation performance by displacing the camera $\pm 2$m along the XY-axis and computing the Fréchet Inception Distance (FID) between the rendered and source images. As reported in \cref{tab:extrapolation}, SaLF and StreetGS achieve comparable extrapolation quality. While NeuRAD demonstrates greater robustness, this is primarily attributable to its CNN post-processing; notably, NeuRAD without its CNN decoder performs similarly to our method.
\paragraph{Efficiency analysis}
Our analysis (\cref{fig:efficiency-tradeoff}) reveals that at lower resolutions ($< 960\times 540$), ray rendering significantly outperforms rasterization due to the latter's resolution-independent projection overhead. As resolution increases, rasterization becomes the more efficient approach, demonstrating the complementary nature and practical value of supporting both rendering paradigms within a single representation. We further analyze the relationship between rendering efficiency and quality, finding a consistent trade-off where higher FPS results in slight PSNR reduction.  More subdivision iterations increase voxel density and detail at the expense of inference speed.

\paragraph{Ablation Study}
{Two key aspects of \methodname\ are its local implicit field representation as a matrix compared  to a fixed scalar value per voxel, and its adaptive voxel pruning and densification during optimization.}
{\cref{tab:ablation_study} reports the camera realism, demonstrating the value of both choices.}

\begin{figure}[t]
	\centering
	\includegraphics[width=1.0\linewidth]{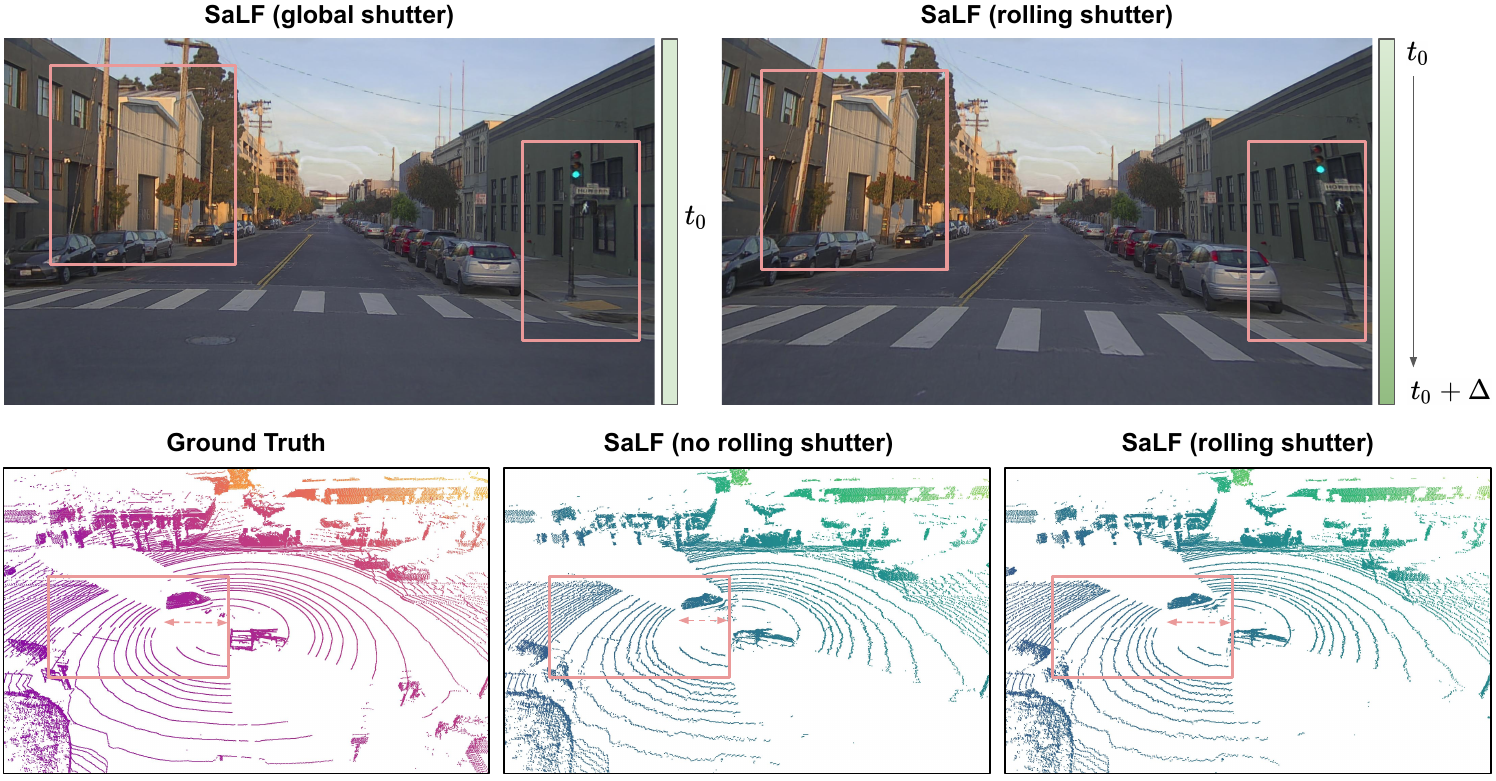}
	\vspace{-0.2in}
	\caption{\textbf{Rolling-shutter simulation via SaLF-Ray.} 
		\textbf{Top:} We render the same view using global shutter and rolling-shutter camera models (see highlighted distorted region).
		\textbf{Bottom:} \methodname{} simulates rolling-shutter effect and accurately match ground truth point clouds (see lidar sweep seam and relative position for the dynamic actor.)}
	\label{fig:rolling_shutter}
\end{figure}

\subsection{Simulation Capabilities and Extensions}
SaLF establishes a foundation for a performant, data-driven multi-sensor renderer that can be readily extended to various downstream simulation needs. As showcased in \cref{fig:teaser}, our method supports ray-based light phenomena (e.g., refraction, shadows of inserted objects) and generalized camera models like panoramic rendering, demonstrating its versatility for diverse sensor rendering tasks. Furthermore, because temporal motion during capture significantly impacts perception~\cite{manivasagam2023towards}, \methodname\ explicitly simulates both LiDAR and camera rolling-shutter effects (\cref{fig:rolling_shutter}). Building on these ray-based capabilities, we also incorporate LiDAR-specific features such as raydrop and intensity simulation (\cref{tab:lid_comparison}) Beyond these, future directions include incorporating beam divergence, actor-label refinement, and sensor pose refinement, offering exciting avenues for the community.

\begin{table}[t]
	\centering
	\caption{{Ablation study on \methodname{} components.}}
	\vspace{-0.1in}
	\begin{tabular}{@{}l|lll@{}}
		\toprule
		\ Models   & PSNR$\uparrow$      & SSIM$\uparrow$     & LPIPS$\downarrow$      \\ \midrule
		\ Ours        &   \textbf{25.48} & \textbf{0.744}  &  \textbf{0.373}      \\
		\  $-$ Densification           &   23.19  &  0.670   &  0.474   \\
		\  $-$ Field matrices              &  25.11 & 0.735 & 0.386  \\
		\bottomrule
	\end{tabular}
	\vspace{-0.05in}
	\label{tab:ablation_study}
\end{table}

\begin{table}[t]
	\centering
	\caption{SaLF extension for LiDAR raydrop and intensity.}
	\vspace{-0.1in}
	\setlength{\tabcolsep}{4pt}
	\begin{tabular}{l|ccc}
		\toprule
		Models & Intensity & Raydrop Acc. & Speed (FPS) \\
		\midrule
		NeuRAD~\cite{tonderski2024neurad} & \textbf{0.062} & \textbf{96.2} & 3.79 \\
		UniSim~\cite{unisim} & 0.085 & 91.0 & 11.8 \\
		SaLF-large & 0.069 & 92.7 & \textbf{350} \\
		\bottomrule
	\end{tabular}
	\label{tab:lid_comparison}
\end{table}

\begin{table}[htbp!]
	\centering
	\caption{{Downstream domain gap evaluation.}}
	\vspace{-0.1in}
	\label{tab:dg}
	\begin{tabular}{lccc}
		\toprule
		& Det. Agg.~$\uparrow$ & Pred. ADE~$\downarrow$ & Plan Cons.~$\downarrow$ \\
		\midrule
		UniSim~\cite{unisim} & 0.74 & 0.63 & 0.99 \\
		Ours & \textbf{0.78} & \textbf{0.52} & \textbf{0.83} \\
		\bottomrule
	\end{tabular}
\end{table}

\begin{table}[htbp!]
\centering
\caption{\textbf{Extrapolation Performance.} FID scores for novel view synthesis when displacing the camera $\pm 2$m along the XY-axis.}
\vspace{-0.1in}
\begin{tabular}{@{}l|c@{}}
\toprule
Models & FID $\downarrow$ \\ \midrule
NeuRAD w/ CNN & \textbf{29.52} \\
NeuRAD w/o CNN & 40.84 \\
Street Gaussian & 37.92 \\
SaLF-large & 41.22 \\ \bottomrule
\end{tabular}
\label{tab:extrapolation}
\end{table}

\subsection{Downstream Applications}
In addition to photometric and geometric realism, the fidelity of a sensor simulator must be evaluated by how accurately it replicates the sensor data as perceived by the downstream autonomy system. We measure the open-loop domain gap on a 187-sequence highway dataset using a modern autonomy stack with a joint detection/prediction transformer~\cite{casas2024detra} and a trajectory planner~\cite{sadat2019jointly}. For each sequence, we run the autonomy on paired real and simulated data, and compare the downstream outputs using three metrics: (1) \textit{detection agreement}: average precision (AP) matching simulated to real detections (IoU $\geq 0.5$); (2) \textit{prediction ADE}: minimum average displacement error between forecasted and ground-truth trajectories; and (3) \textit{planning consistency}: endpoint deviation between 5s plans generated from real versus simulated inputs. As shown in \cref{tab:dg}, SaLF achieves lower domain gap across detection, prediction, and planning, outperforming UniSim on all downstream tasks.

\subsection{Limitations}
Our method typically requires a higher number of voxels compared to {3DGS-based StreetGaussian} to achieve comparable rendering quality. This stems from {our voxels having} fixed size, position, and orientation, rather than adaptive primitives (like 3DGS) that dynamically adjust their shape to local structures.
{We also note that additional modifications are required to support non-rigid and temporal changes in our scene representation \cite{chen2024omnire}.
While our method supports full raytracing, and we demonstrate phenomena such as shadow in \cref{fig:teaser}, we train \methodname\ using primary rays only.}
\vspace{-0.01in}

\section{Conclusion}

\label{sec:conclusion}
In this work, we tackled the problem of developing a multi-sensor simulation system that is fast to train, realistic, and efficient to render with.
Towards this goal, we proposed a novel representation, \methodname, which consists of a set of sparse voxels, and where each voxel defines a local implicit field.
Importantly, we design our representation to support both
rasterization and raycasting,
enabling support of ray-based phenomenon such as rolling-shutter, shadows and refraction, as well as sensors with distorted lenses,
which was previously difficult to achieve with 3DGS.
We enhance \methodname\  for driving scenes via multi-scale voxel initialization, adaptive pruning and densification, and dynamic actor modelling.
We demonstrated that \methodname\ achieves comparable LiDAR and camera realism to existing neural rendering simulation methods,
while being much faster to train (up to 5$\times$) and render with (over 100$\times$ for LiDAR), enabling more scalable sensor simulation for autonomy development.

\section{Acknowledgements}

\label{sec:acknowledgements}
We thank Ioan-Andrei Barsan and Yasasa Abeysirigoonawardena for constructive discussion and feedback. We thank the Waabi team for their valuable assistance and support.

\bibliographystyle{IEEEtran}
\bibliography{references}

\end{document}